\pdfoutput=1
\relax
\documentclass[letterpaper]{article} 
\usepackage{aaai22}  
\usepackage{times}  
\usepackage{helvet}  
\usepackage{courier}  
\usepackage[hyphens]{url}  
\usepackage{graphicx} 
\usepackage{natbib}  
\usepackage{caption} 
\DeclareCaptionStyle{ruled}{labelfont=normalfont,labelsep=colon,strut=off} 
\usepackage{algorithm}
\usepackage{algorithmic}

\usepackage{newfloat}
\usepackage{listings}
\floatstyle{ruled}
\newfloat{listing}{tb}{lst}{}
\floatname{listing}{Listing}
\graphicspath{ {./} }

\usepackage{amsmath}

\title{Designing Language Technologies for Social Good: The Road not Taken}
\author{
    Namrata Mukhija\textsuperscript{\rm 1}, Monojit Choudhury\textsuperscript{\rm 2}, Kalika Bali\textsuperscript{\rm 2}
}
\affiliations{
    \textsuperscript{\rm 1}New York University\textsuperscript{}\thanks{Work done as part of internship at Microsoft Research Lab India}\\
    \textsuperscript{\rm 2}Microsoft Research Lab India
}
\usepackage{bibentry}

\begin{document}

\maketitle

\begin{abstract}
Development of speech and language technology for social good (LT4SG), especially those targeted at the welfare of marginalized communities and speakers of low-resource and under-served languages, has been a prominent theme of research within NLP, Speech and the AI communities. Researchers have mostly relied on their individual expertise, experiences or ad hoc surveys for prioritization of language technologies that provide social good to the end-users. This has been criticized by several scholars who argue that work on LT4SG must include the target linguistic communities during the design and development process. However, none of the LT4SG work and their critiques suggest principled techniques for prioritization of the technologies and methods for inclusion of the end-user during the development cycle. Drawing inspiration from the fields of Economics, Ethics, Psychology and Participatory Design, here we chart out a set of methodologies for prioritizing LT4SG that are aligned with the end-user preferences. We then analyze several LT4SG efforts in light of the proposed methodologies and bring out their hidden assumptions and potential pitfalls. While the current study is limited to language technologies, we believe that the principles and  prioritization techniques highlighted here are applicable more broadly to AI for Social Good.
\end{abstract}

\section{Introduction}
\noindent Language technology plays a vital role in bridging the digital divide by providing natural and intuitive digital interfaces as well as access to information in the user's native language. Several studies, research projects, academic and practice workshops have been conducted under the broad theme of {\em Language Technology for Social Good} (LT4SG). Nevertheless, most LT4SG have been built with little or no engagement with the end-user communities \cite{caselli-etal-2021-guiding, Floridi2020, Fortuna2021CartographyON}. This leads to a multitude of issues such as unanticipated failures of the developed technologies, missed opportunities for social impact, and dual use where technology is re-purposed for negative applications, to name a few. \citeauthor{joshi-etal-2019-unsung} \shortcite{joshi-etal-2019-unsung} remark that the current trajectory for advancement of language technologies caters to the needs of high-resource communities, and has left the vulnerable sections further marginalized due to this digital divide. 

An understanding of individual and community needs is a central pillar to building useful, socially-preferred, sustainable, and democratic language technologies. Any attempt to calibrate technology tools and applications to the needs of a community requires a dialogue with its users to better understand what the end user values in a technology and why. \cite{caselli-etal-2021-guiding}. While limited availability of resources also needs to be factored in for building any technology,  prioritization of building LT4SG on the basis of cost and effort required should also take into consideration users' values and preferences. Ultimately, the two central questions that need to be answered before one ventures into designing and developing LT4SG solutions are:
 (1) Which technology should one develop to maximize social good for a community, given the resource constraints? and (2) In order to meaningfully answer the first question, what information is required and which methodological principles should be followed?

In this paper we introduce a set of methods that can be used to include communities in the design of LT4SG and elucidate needs and preferences of the end-users. We draw diverse ideas from the areas of Economic Theory, Psychology, Ethics, and Participatory Design to propose methodologies that can be used by researchers to collect, analyze, and infer user preferences and thereby design and build language technology in alignment with the end-user needs. To start with, we discuss a set of ground realities at the intersection of user, community, and technology which shape the methodologies described in the paper. This is followed by setting out the challenges faced in inclusive design of LT4SG and principled ways of resolving them through methodologies and an initial stage framework. We critically review three existing studies in machine translation for social good to elucidate the shortcomings in some of the common approaches to building and deploying LT4SG. We propose a methodology that addresses not only these specific drawbacks but also provides a concrete framework for technologists to optimize resource constraints while maximizing social impact on the community.

\section{Related Work}
There is a significantly large and growing body of work on LT4SG; our aim here is not to survey those. The current study is a methodological critique of the existing approaches to LT4SG, and hence we review only those work that have proposed meta-framework for situating and evaluating LT4SG studies, or have raised concerns about the approaches taken. We do this from the lens of the two central questions outlined earlier, around optimizing resources and social good, and the methodologies required to do so.

A common framework for classification of LT4SG work invokes the topical subareas or domains of social good v\`{i}s-a-v\`{i}s resource constraints. This is a prescriptive method, where one already assumes that certain technologies are universally useful. For example, \citeauthor{jin-etal-2021-good} \shortcite{jin-etal-2021-good} provide a research priority list based on the constraints of resources (data procuring, training of researchers etc.) and availability of support (funds, government support, etc.). They draw an Important/Neglected/Tractable (INT) framework and estimate the cost-effectiveness of contributing a unit time and effort of a certain researcher or team to research on a particular technology. 

While resource constraints and availability of support are a reality, any prioritization list of LT4SG must involve end-users in determining it, and include end-user preferences in estimating the cost effectiveness of certain technology. \citeauthor{joshi-etal-2019-unsung} \shortcite{joshi-etal-2019-unsung} and \citeauthor{bird-2020-decolonising} \shortcite{bird-2020-decolonising} present a strong critique of current user-oblivious approach to LT4SG, and provide convincing arguments in favor of ``technology for us - not without us" paradigm. \citeauthor{joshi-etal-2019-unsung} \shortcite{joshi-etal-2019-unsung} argue that the true benefit of LT4SG should not be measured in terms of the accuracy of the systems, but in measurable socio-economic benefits the system brings to target users. Therefore, they propose, successful deployments of language technologies in low resource context should involve a launch by seeding with target communities, working closely to engage the community itself with the technology and information, and provide a strong incentive structure to the target community to adopt the technology. Similarly, \citeauthor{bird-2020-decolonising} \shortcite{bird-2020-decolonising} recognizes the need for cross culture encounters and recommends identifying a recognition space which is culturally safe, where indigenous and external actors can identify common goals and tasks: ``How do speech and language technologies serve to expand people’s capability to lead the kind of lives they have reason to value?".

In fact, the need of community participation in designing LT4SG has been emphasized by several other lines of work. \citeauthor{Cowls2021} \shortcite{Cowls2021} argue that AI systems can cause unanticipated failures if they are not shaped by human values. The authors also caution that there may be countless examples of missed opportunities  to exploit the benefits of AI-based interventions, especially when these systems are built in isolation from the users that are directly impacted by them. 

Tomasev et al. (2020) details the importance of collaboration and deep partnerships in building AI4SG systems that achieve a sustained impact. Similarily, Fortuna et al. (2021) propose that LT4SG needs collaborations from users, activists, minorities, grassroots movements, businesses, non-governmental organizations (NGOs),and social entrepreneurs to achieve a positive societal impact. The authors also mention that when discussing definitions of social good, "positive impact" depends on the context and set of values for the target community.

The above works, while encouraging end-user participation, do not describe any systematic approach(es) for including the end-user in the development cycle of LT4SG.

\section{Laying the Foundation}
We begin by defining a few terms.
\textit{User} or {\em end-user} of a language technology is a speaker of that language who applies the language technology to fulfil their need(s). A \textit{community} is a set of users who share a common language (more specifically a \textit{linguistic community}) and/or one or more of certain other demographic aspects such as age, gender, socio-economic status and education level. The \textit{utility} of a language technology is defined as the satisfaction achieved by an {\em end-user} on applying a language technology to fulfil a need or set of needs. The \textit{benefit} of a language technology is the realized improvement in well-being and quality of life of an individual on applying that technology. Finally, the \textit{Cost} of a language technology is the cumulative project funding required for development and deployment of the technology.

The three entities -- the user, the community and the language technology -- interact with and influence one another in complex ways. It is necessary to understand and model these interactions if we are to systematically investigate the two central questions of LT4SG. Fortunately, these aspects are well-studied by economists, sociologists, philosophers, psychologists and design researchers, which we shall build our foundations on. 
Figure 1 presents an abstract representation of the 3 way bidirectional interactions between these three parties in the form of a triangle. We explain these various interactions and their theoretical foundations below.

{\bf User $\to$ Technology:} Users or speakers of a language know best what utility a technology brings to them. This assumption is grounded in the Rational Choice Theory, propounded by \citeauthor{Smith+2021} \shortcite{Smith+2021}. According to this theory, people don't randomly choose from a range of options but rather have a logical decision-making process that is aligned with their self-interest and takes into account the costs and benefits of various options. In the context of language technology, this idea has been argued for most notably by \citeauthor{bird-2020-decolonising} \shortcite{bird-2020-decolonising} and  \citeauthor{joshi-etal-2019-unsung} \shortcite{joshi-etal-2019-unsung}.

\begin{figure}[t]
\includegraphics{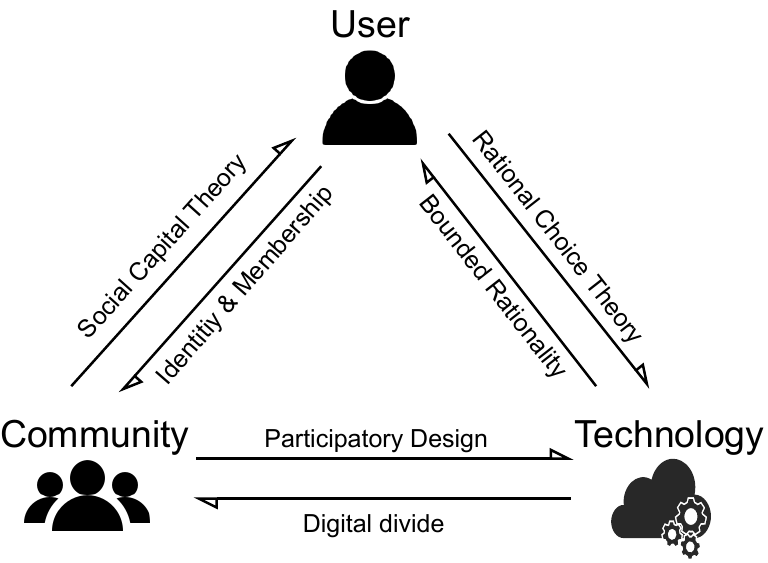}
\caption{Triangular representation of the 3 way bidirectional interactions between \textit{user}, \textit{community}, and \textit{technology}..}
\end{figure}

{\bf Technology $\to$ User:} While Rational Choice theory helps us view users as rational decision-makers, this rationality is bounded by the reality of human cognition \cite{simon1957models}. Limitations include the difficulty of the problem requiring a decision, cost of gathering and processing information, the cognitive capability of the mind, and the time available to make the decision.
Therefore, bounded rationality reassesses the idea of "perfect rationality" as perfection in rational decisions is practically hard to achieve due to the complexity of natural decision problems and the limited computational resources available. 
Since access to digital technology in today's world is a key determinant of access to information, technology itself could be seen as a limiting or influencing factor on a user's rationality in making choice over technology. This creates a potentially vicious cycle of influence, which is important to break or at the least accounted for while designing LT4SG. For instance, \citeauthor{nekoto-etal-2020-participatory} \shortcite{nekoto-etal-2020-participatory} identified that many stakeholders in the process of low-resource Machine Translation were missing invaluable language and societal knowledge, or the necessary technical resources, knowledge, connections, and incentives to form interactions with other stakeholders in the process. Due to limited knowledge and experience of technology, individuals may not consider the full range of costs and benefits while choosing an optimal language technology but rather, choose an option that fulfils their adequacy criteria  \cite{Campitelli2010HerbertSD}.

{\bf Community $\to$ User:} Users are embedded in a community (or within several overlapping communities), and language -- the vehicle for communication and collaboration  -- is inherently a social construct. Therefore, when it comes to adoption of LT, a user's ``rational choices" are often influenced by other members of the community with whom the user has a trust relationship. Referred to as social capital theory   \cite{coleman1988social}, this provides an important lens for looking at LT4SG design, where the technologist must understand the social structures within which the technology operates. As the primary function of language is to transmit life-crucial knowledge \cite{joshi-etal-2019-unsung, bird-2020-decolonising},  users of a language will collaborate with their trusted social relationships to get and combine resources which then can be used to solve social problems \cite{BRIONES201137,  https://doi.org/10.1111/j.1083-6101.2012.01576.x}.

{\bf Community $\to$ Technology:} This indeed is used utilized in the Participatory Design (PD) paradigm of technology design, where all stakeholders (end-users, researchers, institutions like NGOs etc.) are invited to collaboratively design the technology. PD is necessary to ensure all voices are heard and interactions essential to developing a sustainable, democratic, and user-preference aligned language technology are facilitated \cite{b41737e94e014bcca571757462e47712}. The members can share and communicate information, build relational and logistical support, and thereby generate solutions for challenging social problems \cite{Barak2020ThePA}. 

{\bf User $\to$ Community:} Language is a crucial part of one's identity; this also means that an individual might not want to associate with a certain linguistic community or identity, even though a technologist might wrongly assume so. While designing LT4SG, one therefore, must be mindful of individual choice of linguistic identity. For instance, a speaker of a minority language might aspire to learn or use technology in a majority language due to socioeconomic opportunities. In this case, it will not only be futile but also overreaching for a technologist to push an LT4SG in the minority language on this user. 
 
{\bf Technology $\to$ Community:} Different technologies cater differently to different communities, creating a sharp digital divide. \citeauthor{joshi-etal-2019-unsung} \shortcite{joshi-etal-2019-unsung} argue how large data-driven NLP technologies have widened the gap between languages that have access to large resources and those which don't. A global perspective on LT4SG must take into account these disparities between languages and linguistic communities, and factor in the aspects of fairness across communities while taking large scale investment and design decisions. \citeauthor{Choudhury2021HowLF} \shortcite{Choudhury2021HowLF} argue that instead of optimizing for average utility across communities, LT4SG could adopt a prioritarian or Rawlsian principle of fairness \cite{41266156}, where the most marginalized communities or languages are to be given highest priority, because it leads to a more egalitarian outcome in the future when all languages eventually develop access to technology and resources. Of course, there are several other principles of distributive justice, and different principles may be more appropriate for different contexts. 

\section{Challenges}
In this section, we present the challenges faced during the prioritization of LT4SG.

\textbf{Preferences are hard to elicit.} While we would like to ask users about their preferences and values, and design technologies that align with those preferences and values, this is not a straightforward process. As discussed in the previous section, users may not have access to the information required to make informed choices between technologies that satisfy their needs, which is often the case for low-resource communities \cite{joshi-etal-2019-unsung}. They may also not have the incentive or means to gather and process this information. Moreover, even if they do have information to make these choices, they may be limited by their cognitive, memory, and attention capabilities in imagining and coming up with hypothetical language technologies which might be beneficial for their community. An additional aspect to consider is, if users are able to communicate their values, these values must be carefully selected and fostered to ensure that societal ills, like existing inequalities and environment problems, are not widened with the use of language technologies \cite{Cowls2021}. 

\textbf{Feasibility of the solution.} There is a cost associated with building any language technology. There is funding involved in various stages of artifact generation in the development cycle of a language technology. Data creation, collection and standardization, model building, infrastructure, and deployment, all require resources, both financial and human. Keeping this cost in mind is important when trading off different technologies to ensure feasibility of selected solutions.

\textbf{Open-ended design space.} Another challenge is the wide and open-ended space of technology. A particular language technology may have many features or attributes that can be built into it. They may have various functionalities that could aid the user in applying that technology towards fulfilling their goals. However, deciding which attribute to include and which to exclude is not a trivial task since the set of attributes is humongous and often trade-offs are required. 

\section{Methodologies}
Given the challenges, we now look at possible solutions, drawn from Economic Theory, Psychology and Participatory Design that can help resolve the two central questions of LT4SG in a principled way.

\subsection{Technology prioritization}
Any prioritization of LT4SG must include end-users in its creation and must be aligned with their preferences to ensure the social impact of the language technologies developed. A meaningful way to ascertain this would be to first elicit user preferences for language technologies that align with their needs and goals. To do so, we would like to quantify {\em willingness-to-pay} (WTP) for a language technology by an individual which would express the maximum utility that an individual gains by using a particular technology. A range of stated preference techniques have been developed for eliciting consumers preferences and allow us to quantify the individuals’ economic valuation or WTP for goods and services. All the techniques involve asking respondents to consider one or more hypothetical options and to express their preferences for them through surveys. They have been extensively applied across various industries including transportation \cite{CHENG20191, WANG2017210}, policy \cite{AZAROVA20191176, RePEc:oup:qjecon:v:133:y:2018:i:1:p:457-507., CHUNG2017289}, farming \cite{KHATRICHHETRI2017184}, and healthcare \cite{article, Zanolini2018UnderstandingPF} among others. There are two main classifications for these techniques: {\em contingent valuation} and {\em multi-attribute valuation}.

\textbf{Contingent Valuation}
In order to determine an individual's WTP for a particular language technology, one way is to break down the system into a bunch of relevant features, and then ask the user whether and how much they are willing to pay for each feature. This way of ascertaining WTP is known as Contingent Valuation (CV) \cite{10.2307/1232747}. For instance, one could ask the individuals in a target linguistic community what is their maximum willingness to pay for enabling voice in an existing text-based search app. However, open-ended questions like these are associated with cognitive load on the users and too difficult to answer because users are not accustomed to paying for non-market goods and services. They may also feel that the described change is infeasible and might be expressing that instead of their true WTP. 

A potential way to ensure that users express their true WTP is to provide them with more information before they take the survey. In fact, a common way to do this across stated preference techniques is by also including all the stakeholders of the targeted linguistic community in discussing the feasibility of the change given the existing environment and infrastructure \cite{KHATRICHHETRI2017184}. We believe that in the context of LT4SG, engagement of stakeholders and the community during the training phase is especially important because on one hand users might feel that the technology is infeasible in their context, and on the other hand, they might overestimate the power of technology and be unaware that in practice all LT will fail to produce the desired output at least in some situations. 

While community engagement might help with the above issues, open-ended questions are still subject to other biases, for example, users ignoring income constraints \cite{10.1257/jep.8.4.45}. Dichotomous CV, where users have to answer in yes-no, imposes less cognitive load due to limited choices in response but results seem to be more positively inflated than open-ended values, possibly due to ``yeah saying" (Hanley et al. 2001). Note that CV helps in eliciting the WTP for only one feature at a time.

\textbf{Multi-Attribute Valuation}
Usually a language technology has multiple features instead of one feature and all these features are valued differently by the users of that language technology. In economics, {\em Theory of value} \cite{Lancaster1966ANA} assumes that consumers' utilities for goods can be decomposed into utilities for its attributes. To understand which language technology to build next for a linguistic community, technologists should understand the value the users in that linguistic community attach to the features. This can be done via a well-known set of techniques from Economics called Multi-attribute valuation (MAV). \cite{RePEc:upf:upfgen:705}. The broad idea behind these techniques is as follows. The user is shown a set of language technologies described by their attributes possibly including the price. The user is then asked to rank, rate or choose between the technologies.From their choice, the utility or WTP of the technologies can be ascertained along with the relative importance of the attributes. MAV techniques are further classified into conjoint analysis and choice modeling. Conjoint Analysis \cite{Luce1964SimultaneousCM, Green1971ConjointMF} is a preference-based approach that uses a deterministic utility function, that is, it doesn't account for any source of uncertainty in the decision-making. Conjoint analysis has two types of techniques nested in it: {\em Conjoint Rating} and {\em Paired Comparison}.

In conjoint rating, all products are presented to the respondent one at a time, with all the attributes and their levels for that product being shown. The respondent then marks their preference on a numeric or semantic scale defined by the researcher. For example, imagine that a respondent is presented with two products, a spell checker and a voice-enabled search, one at a time. The spell checker is built in the respondent's native language which is low-resource while the voice-enabled search is built in a high-resource language, say English. The spell checker requires a keyboard interface to interact with it and the voice-enabled search requires a microphone to interact with it. Also, one obvious attribute is that the spell checker is text-based whereas the voice-enabled search is voice-based. Additionally, the spell checker costs \$100 and the voice-enabled search costs \$30 due to the higher cost of data acquisition in the respondent's native (low-resource) language. The respondent is then presented with these two product profiles (including the language and price attributes for each of the product) and the respondent is asked to rate each of them on a scale of 0 (least preferred) to 10 (most preferred). 

As seen, this method does not involve a direct comparison of alternative choices and ratings must be transformed into a utility scale. It has the implicit assumption of cardinality of rating scales and the implicit assumption of comparability of ratings across individuals, both of which are inconsistent with consumer theory \cite{RePEc:upf:upfgen:705}. Therefore, preference-based methods were proposed \cite{Louviere1988AnalyzingDM}, that focus more on gaining an insight into consumer preferences rather than estimating economic values. In paired comparison \cite{Thurstone1927-THUALO-2}, a preference based method, respondents are asked to choose their preferred alternative out of a set of two choices and to indicate the strength of their preference in a numeric or semantic scale. Direct comparison of the two choices or product profiles presented is involved.

Both the techniques under Conjoint Analysis, due to the use of a deterministic utility function, do not account for the effects of incomplete knowledge on the choices a particular user make. It also fails to incorporate randomness in choice which may happen if there is no preference among the presented alternatives. This may lead to wrong conclusions about which language technologies and attributes are valued by the community. To mitigate this, the presented alternatives must have a status quo option that has the language technology with same attribute levels that the community has been leveraging to satisfy their needs and are well-acquainted with.

\textbf{Choice Modeling:} In real life, users will compare multiple competing language technologies before choosing one. They might also have a ranking of those language technologies from most preferred to least preferred. Eliciting this full ranking for each user will provide additional information for estimating the preferences across a community. Technologists should account for incomplete knowledge that they have about the community and also factor into their decision-making, incomplete knowledge among the community members and the irrational behavior resulting from them (for example, the user not choosing the alternative providing them the maximum utility). The more natural choice setting this requires is formalized in choice modeling, or choice-based approaches. Choice-based approaches are based on a real-life task that consumers perform every day, that is, the task of choosing a product from among a group of competitors. In contrast, preference-based approaches do not require respondents to make a commitment to select a particular option \cite{RePEc:upf:upfgen:705}. The products are described as a series of attributes along with the levels each attribute takes, same as Conjoint Analysis. The underlying foundation is Random Utility Theory \cite{Domencich1977URBANTD} which assumes that the utility of each alternative is the sum of a deterministic portion and an unobserved or random portion. Different discrete choice models are obtained from different specifications of the random portion of utility.  Another underlying foundation is that the choice behavior underlying each choice satisfies Luce’s \shortcite{Luce59} choice axiom, well known as the independence from irrelevant alternatives (IIA). The independence from irrelevant alternatives means that the random portion of utility for one alternative is unrelated to the random portion of utility for another alternative.

We next consider the two different kinds of experiments in choice modelling: single-choice experiment and ranking experiment. In single-choice experiments \cite{train_2009}, respondents are presented with a series of alternatives and asked to choose their most preferred option. If we assume that the unknown portion of utility for each individual is independent and identical, one can use Multinomial Logit Model (MNL) (McFadden 1973) to estimate the coefficients of the attributes. In ranking experiment \cite{Beggs1981AssessingTP, Chapaaan1982ExploitingRO}, respondents are asked to rank their options from the most to the least preferred. A ranking of set of alternatives is equivalent to the following sequence: the highest-ranked alternative is chosen over all the other alternatives, the second-ranked alternative is chosen over all alternatives except the first, and so on. The motivation for using this method is to obtain more information for a given data set than the single-choice method can provide. The single-choice process can be seen as a special case of the ranking process, in which respondents simply choose the best alternative.

One of the drawbacks of rank-ordered models is random ranking for lower positions due to fatigue from too many alternatives to rank, or lack of preference between the remaining (less-preferred) alternatives \cite{Dijk2007ARL}. A solution is working with partial rank-ordered data and consider only the top $k$ ranks as illustrated in \citeauthor{Chapaaan1982ExploitingRO} \shortcite{Chapaaan1982ExploitingRO}. Logit models cannot capture preference variations that happen with respect to unknown attributes or are purely random \cite{train_2009}. Therefore, if there are attributes of the language technologies presented that are not elicited but influence the choices of the individuals in the community then the conclusions from the estimates of a logit model will be misleading. Another thing to consider is the substitution pattern enforced by the IIA assumption in the logit model. If the attributes of a language technology improve (for example, the cost changes) then individual might prefer this language technology more or less over others. An increase (or decrease) in probability of choosing this alternative will have necessarily mean a decrease (or increase) in probability for other alternatives since the choice probabilities sum up to one. While this substitution pattern might be appropriate in some contexts, it is not in others \cite{10.2307/1907717, Ortzar1983NestedLM, Brownstone1998ForecastingNP, debreu1960individual}. Probit models \cite{Thurstone1927-THUALO-2} which are derived under the assumption of jointly normal unobserved utility components allow for dealing with the above issues. The normal distribution for the representing random components may not be ideal for certain situations. For example, for price coefficients, a normal distribution would imply a positive coefficient for price for some individuals which is not the case usually. To overcome this limitation, the researcher can use a mixed logit model which can approximate any random utility model \cite{McFadden2000MIXEDMM}. 

In MAV techniques, deciding the attributes of a product and the choice of products itself should be expected to influence preferences, clear and communicable, and be relevant to decision-making. We propose a set of methodologies in the next section for eliciting attributes that are important to the user.

\subsection{User Knowledge and Values Elicitation}
Attributes a consumer values of a product is believed to be determined by the consumer's intrinsic and extrinsic  motivation \cite{https://doi.org/10.1111/j.1467-6435.1991.tb01759.x}. The intrinsic motivation linked to a certain attribute is determined by the consumer's perception of the instrumentality of the attribute as regards the satisfaction of the consumer's needs, i.e.,the extent to which the attribute is inherently linked to consumer benefits. The extrinsic motivation is associated with the consumer's perception of whether the set of products available and acceptable to the consumer,i.e.`the choice set' differs with regard to the attribute in question (Bech-Larsen and Nielsen 1999).
A very wide range of elicitation techniques have
been proposed in literature, grounded in the field of Psychology, to elicit attributes and more generally, knowledge: interviews \cite{Johnson1987}, protocol analysis \cite{ericsson1984protocol},
laddering \cite{10.1016/S0020-7373(87)80094-9}, work groups \cite{wood1995joint}, triadic sorting \cite{george1955psychology} and focus groups \cite{87999} to name a few. 
We would also like to understand the intangible personal and emotional values that drive the decision-making of the end-user. This can be done through Means-End Chain theory which is a value-based, cognitive model \cite{reynolds2001understanding}. It involves a three-step process: identification of the salient product attributes, the laddering procedure and analysis of the data, and plotting of the Hierarchical Value Map (HVM). The resulting HVM consists of nodes, which represent attributes, consequences, and values, and lines connecting these nodes, which represent the frequency of linkages between them. The above techniques can help researchers have a better understanding of the user's mental model and how much information is available with them to make informed decisions.

\begin{figure}[t]
\includegraphics{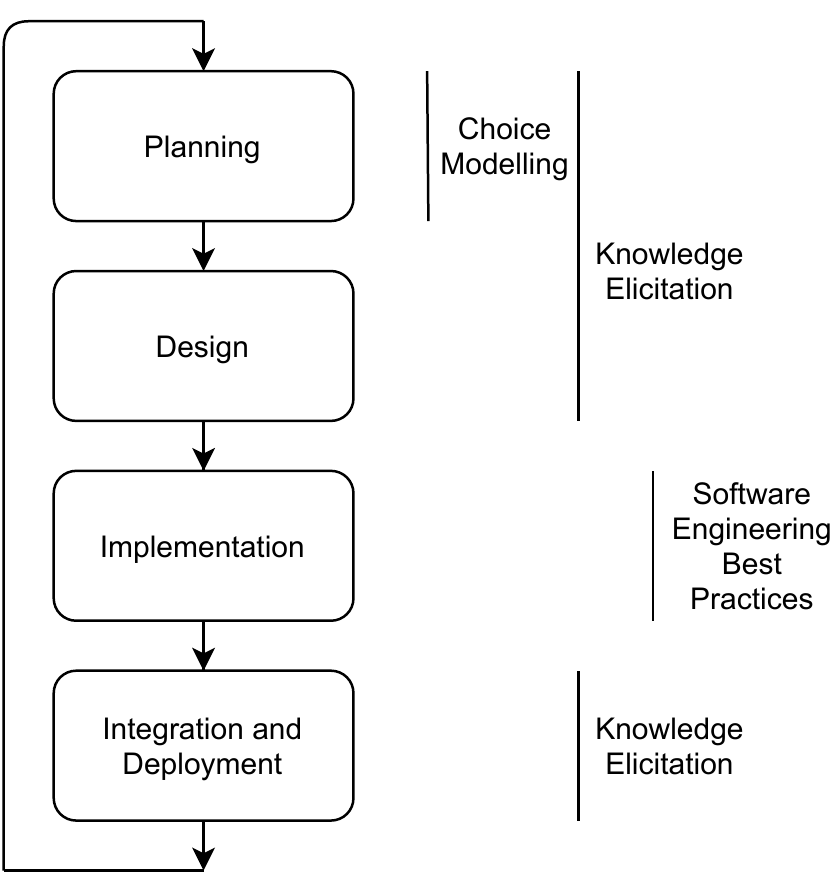}
\caption{Lifecycle representation of NLP4SG development showing methodologies applicable in each stage.}
\end{figure}

\section{Framework}
We now present the framework (Figure 2) to answer the two central questions of LT4SG using the previously defined methodologies. The framework proposed here is an initial formation of how the methodologies presented are involved in different stages of NLP4SG lifecycle. There are multiple works which have highlighted the development cycle for building a product and we use them to highlight different areas where (as well as how) we can involve communities. Most of the work can be done in the planning and design phase and the information gained about the community values and preferences can be utilised in the following stages. 

In the planning and design phase, it is important to understand what the end-user values in terms of language technology products and their attributes. For example, suppose safety is a value that is prioritized by members of a community. Being connected and being able to convey important information to the other members of the community makes them feel safe. Does a voice-based product facilitates communication better than a spell checker for the community under consideration? Attribute elicitation techniques mentioned in the previous section can be useful in answering this question. Another way to facilitate communication of needs between end-users and technologists is the use of knowledge elicitation techniques like card sorting and laddering. Through card sorting the open-ended design space can be navigated with structure and can help reach a common understanding of how the user perceives language technology products. Laddering can uncover values (like safety) and consequences (like being connected) which drive users to prefer certain attributes (voice over spell checking or vice-versa) over others. The technologist with other necessary agents can then make appropriate judgment on which values should be fostered and determine which  attributes to incorporate. Once knowledge and values of the users are elicited, the technologist has good understanding of user's mental model and the user also has a shared understanding of the design space with the technologist. This an iterative phase and may require multiple such engagements to get a holistic understanding.
With the technology products and their attributes that have come up in the planning phase, the technologist can determine the resources to build those hypothetical products and assign costs to them. Next, the technologist can conduct a full-scale choice experiment to understand preferences for language technology products and determine the WTP for different products. Influence of government support and other policy-making bodies can be ascertained in this stage.

Once prioritization of language technology from the user perspective is known, those language technologies can be developed. The attributes elicited by the end-users in the planning and design phase should be incorporated. In the development phase, if the target community is linguistically diverse, Rawlsian principle of fairness \cite{41266156}, as argued by \citeauthor{Choudhury2021HowLF} \shortcite{Choudhury2021HowLF}, may be used to choose among various multi-lingual models developed. There are multiple ML life-cycle works that highlight best practices for this stage, and we defer the reader to them \cite{serban2020adoption, de2019understanding}. 

During integration and deployment, the steps suggested by \citeauthor{joshi-etal-2019-unsung} \shortcite{joshi-etal-2019-unsung} can be followed. Seeding can be done with the group of users that participated in the knowledge elicitation stage and incentives for usage and sharing can also be decided upon in the knowledge elicitation stage. Also, if governmental (or other agencies) support had huge influence in the choice elicitation stage, then the technologists can work with the those agencies to encourage usage among users. 

\section{Case Studies}
We look at very recent papers which have come closest to using some of the ways we have described in this paper and have been successful in their respective pursuits. We have arranged the case studies such that the interactions between user, community, and technology as shown in Figure 1 are stronger progressively.

\textbf{Case Study 1.}
TICO-19: the Translation Initiative for COvid-19 is a collaborative work done by Carnegie Mellon University, Translated, Amazon AI, Microsoft, Facebook AI, John Hopkins University, Appen, Google and Translators without Borders to enable translation of content related to COVID-19 into a wide range of languages \cite{anastasopoulos2020tico}. This work showcases language technology in action, aiding to save lives during the worst pandemic to hit the world in a century. They identify the need for multilingual communication to exchange important and correct information with at-risk populations in emergencies and provide multiple translation artifacts which can help accelerate multilingual machine translation tremendously. They create a TICO-19 benchmark that translates COVID-19 source data into 35 languages by first sending each document for translation to a language service provider (LSP) and then perform post editing on the translation by qualified professionals familiar with medical domain. While usage of LSPs is common in literature, there are chances of mistranslation which can further exacerbate misinformation, specially given that how these LSP are chosen is not described in the paper. Making sure that the LSP chosen are representative of the target language community can help mitigate possibilities of mistranslations. They interact with Translators without Borders (TWB), an intermediary, who has experience and context dealing with situations that require urgent communication in multiple languages. This is a good proxy to interacting with the end-user and we assume that TWB understands the needs of its users well. However, collaboration with the end-user through PD methods highlighted in the previous section can be conducted and are essential in understanding if the artifacts generated actually make a tangible impact in improving the experience of the targetted communities. Further, in the analysis of the baselines they observe high disparity in performance for high-resource and low-resource language. A Rawslian principle of fairness can be adopted where the multilingual model chosen minimizes the maximum difference between the BLEU scores for high-resource and low-resource languages to narrow this disparity.

\textbf{Case Study 2.} \citeauthor{santy2021language} \shortcite{santy2021language} investigated mixed-initiative approaches to address some of the social challenges of machine translation. We chose this work as it demonstrates how language technology can underwhelm some of the pain points that users face in their everyday tasks. They introduced a new interface called INMT to a non-profit organization (NGO) Pratham Books that publishes quality books for children in 280 languages. This interface provided suggestions to community translators in two forms: a full-sentence gisting and two-word drop down suggestions. These interactive suggestions were aimed to reduce translation time, increase quality, prime the translator, and reduce the drafting requirements. In this work as well the authors work with an intermediary to provide a solution and then conduct feedback user studies to understand if the system actually works, which is a practical way to develop language technologies. However, they only look at the situation from a technologist lens and focus only on the language aspect of the problem which ignores surrounding context of user setting like availability of low-bandwidth to even use the internet-powered interactive interface properly. We also see an underlying assumption from the technologist's end about the needs of the community translators which was brought to light in their user study as well. The community translators preferred that the translation happened within the story paragraphs instead of sentence-wise. They also preferred that the creativity in the suggestions be fine-tuned as per the story level. These preferences, as evident from the feedback they received, originated from the need of the translators to be aided in the process of translation instead of automating the process altogether. The approaches highlighted for PD and eliciting preferences both can help in facilitating a dialogue between the researchers and community translators at Pratham Books to ensure preferences and needs are well understood a prior to design. Moreover, even if we choose to frame the problem space through only a technologists perspective, while designing this interface they faced challenges that frequently result from an open-ended design space. One of them was providing mouse functionality or a keyboard functionality for the users to interact with this system. They observed different types of devices used by translators at Pratham Books highly influenced which functionality they preferred. Another one was providing suggestions throughout the translation process or only when needed. They mention in their feedback user studies that providing suggestions throughout the translation process tend to throw off the translator and interfere with their thought process. If the design of the interface is done using PD techniques and with the user, it can result in saving iterations and repeated effort in the development process.

\textbf{Case Study 3.}
Masakhane is a grassroots organisation whose mission is to strengthen and spur NLP research in African languages, for Africans, by Africans. They have extremely successful in doing so and have connected agents involved in language technology across the world. \citeauthor{nekoto-etal-2020-participatory} \shortcite{nekoto-etal-2020-participatory} used participatory research to identify and involve all necessary agents required in the Machine Translation development process. They identified missing interactions between content creators and data curators leading to noisy translation pairs, and between stakeholders and evaluators leading to unsuitable evaluation metrics. They facilitated the community to forge these missing connections online by using Github and Slack as the medium of interaction and provided incentive to participate in the form of mentoring. A unique point of this initiative is that it is a technologist-drive effort but those technologists are also end-users and consume the artifacts they generate. Thus, the design of the language technology happens with the people, and for the people of the target linguistic community. They report multiple research outcomes in the form of research artifacts, dataset creation, and published benchmarks. While the success of this initiative is unparalleled, to be principled in goal setting, an initial back and forth dialogue between the researchers and communities using the methodologies in Section 5 to shape research goals can be fostered. Participants of low-resource communities may not understand the benefits of language technology or be able to identify and make informed choices on which technology will serve their needs best due to lack of information as we highlighted in Section 4. Therefore, they may not be motivated to participate. To exemplify the social impact of this initiative beyond the scope of language it already caters to, it is imperative to let the burden of initiative remain with the researchers. PD methods can help in reaching out to more communities and setting context for future collaborations. Additionally, since the interactions require internet and constant connectivity, offline PD methods can help the initiative be more accessible.  
\section{Conclusion}
We proposed principled ways of prioritizing LT4SG research and eliciting preferences, needs, knowledge, and values of a user. We looked at certain ground realities which shape our proposed methodologies as well challenges in involving communities during the development cycle of LT4SG. We also highlight hidden assumptions and propose remedies in three recent and relevant efforts in LT4SG. Community engagement is a central tenant in developing LT4SG and we hope future work can utilize the methodologies highlighted in this paper to ensure that language technology is built with linguistic communities, for linguistic communities, and by linguistic communities.

\bibliographystyle{bibtex}
\bibliography{mukhija} 


\end{document}